\documentclass[11pt,a4paper]{article}
\usepackage[hyperref]{emnlp-ijcnlp-2019}
\usepackage{times}
\usepackage{latexsym}

\usepackage{xcolor}
\usepackage{amsmath}

\usepackage{tabularx,pbox}
\usepackage{times}
\usepackage{latexsym}
\usepackage{amsmath}
\usepackage{amsfonts}
\usepackage{color}
\usepackage{graphicx}
\usepackage{booktabs}

\usepackage{url}

\usepackage{rotating}
\usepackage{multirow}

\aclfinalcopy 

\title{Translate and Label! An Encoder-Decoder Approach for \newline
Cross-lingual Semantic Role Labeling}

\author{Angel Daza and Anette Frank \\
   Leibniz ScienceCampus ``Empirical Linguistics and Computational Language Modeling''\\
   Department of Computational Linguistics\\
   Heidelberg University \\
   69120 Heidelberg, Germany\\
  {\tt \{daza,frank\}@cl.uni-heidelberg.de}
 \\}

\date{}

\begin{document}
\maketitle
\begin{abstract}
We propose a Cross-lingual Encoder-Decoder model that simultaneously translates and generates sentences with Semantic Role Labeling annotations in a resource-poor target language. Unlike annotation projection techniques, our model does not need parallel data during inference time. Our approach can be applied in monolingual, multilingual and cross-lingual settings and is able to produce dependency-based and span-based SRL annotations. We benchmark the labeling performance of our model in different monolingual and multilingual settings using well-known SRL datasets. We then train our model in a cross-lingual setting to generate new SRL labeled data. Finally, we measure the effectiveness of our method by using the generated data to augment the training basis for resource-poor languages and perform manual evaluation to show that it produces high-quality sentences and assigns accurate semantic role annotations. Our proposed architecture offers a flexible method for leveraging SRL data in multiple languages.

\end{abstract}

\normalsize

\section{Introduction}

Semantic Role Labeling (SRL) extracts semantic predicate-argument structure from sentences. This has proven to be useful in Neural Machine Translation (NMT)
\cite{marcheggiani18-Intro}, Multi-document-summarization \cite{Khan15-Intro}, AMR parsing \cite{Wang15-Intro-AMR} and Reading Comprehension \cite{mihaylovfrank:2019}. SRL consists of three steps: i) predicate detection, ii) argument identification and iii) role classification. In this work we focus on PropBank SRL \cite{Palmer05-SRL}, which has proven its validity across languages \cite{vanDerPlas10-Proj}. While former SRL systems rely on syntactic features \cite{Punyakanok08-SRL, Tackstrom15-SRL}, recent neural approaches learn to model both argument detection and role classification given a predicate \cite{March17-SRL, He17-SRL}, and even jointly predict predicates inside sentences \cite{He18-SRL, Cai18-SRL}. 
While these approaches alleviate the need for pipeline models, they require sufficient amounts of training data to perform adequately. To date, such models have been tested primarily for English, which offers a considerable amount of high-quality training data compared to other languages. The lack of sufficiently large SRL datasets makes it hard to straightforwardly apply the same architectures to other languages and calls for methods to augment the training data in lower-resource languages.

\begin{figure}
\centering
  \includegraphics[width=0.49\textwidth]{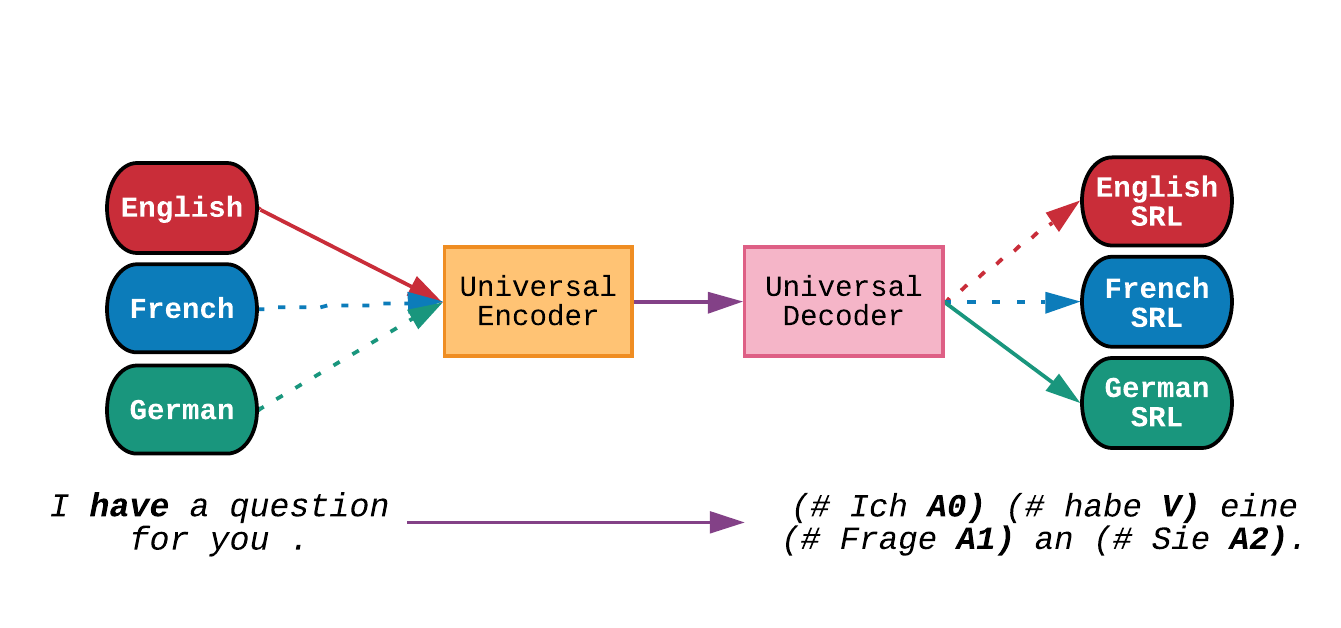}
  \caption{We propose an Encoder-Decoder model that translates a sentence into a target language and applies SRL labeling to the translated words. In this example we translate from English to German and label roles for the predicate \textit{have}.
  }
\label{fig:model_intro}
\end{figure}

There is significant prior work on SRL data augmentation \cite{hartmann17-assessing}, annotation projection for monolingual \cite{furstenau-12-Proj, hartmann-16-gen}, and cross-lingual SRL \cite{Pado09-Proj,vanDerPlas11-Proj,Akbik15-Proj,Akbik16-Proj}. A drawback of cross-lingual projection is that even at prediction time it requires parallel sentences, a semantic role labeler on the source side, as well as syntactic information for both language sides. Thus, it is desirable to design an architecture that can make use of existing annotations in more than one language and that learns to translate input sentences to another language while transferring semantic role annotations from the source to the target. 

Techniques for low-resource Neural Machine Translation (NMT) show the positive impact on target predictions by adding more than one language during training, such as Multi-source NMT \cite{Zoph16-NMT} and Multilingual NMT \cite{Johnson16-NMT, Firat16-multiNMT}, whereas \citet{Mulcaire18-SRL} show the advantages of training a single polyglot SRL system that improves over monolingual baselines in lower-resource settings. In this work, we propose a general Encoder-Decoder (Enc-Dec) architecture for SRL (see Figure \ref{fig:model_intro}). We extend our previous Enc-Dec approach for SRL \cite{Daza18-SRL} to a cross-lingual model that translates sentences from a source language to a (lower-resource) target language, and during decoding jointly labels it with SRL annotations.\footnote{Code is available at: \url{https://github.com/Heidelberg-NLP/SRL-S2S}.}

Our contributions are as follows:\\[-5mm]

\begin{itemize}
    \item We propose the first cross-lingual multilingual Enc-Dec model for PropBank SRL. \\[-7mm]
    \item We show that our cross-lingual model can generate new labeled sentences in a target language without the need of explicit syntactic or semantic annotations at inference time.\\[-7mm] 
    \item Cross-lingual evaluation against a labeled gold standard achieves good performance, comparable to monolingual SRL results.\\[-7mm]
    \item Augmenting the training set of a lower-resource language with sentences generated by the cross-lingual model achieves improved F1 scores on the benchmark dataset. \\[-7mm]
    
    \item Our universal Enc-Dec model lends itself to monolingual, multilingual and crosslingual SRL and yields competitive performance.\\[-4mm]

\end{itemize}

\noindent

\section{An Extensible 
Model for SRL}

\subsection{One Model to Treat Them All}

\begin{figure*}
\centering
  \includegraphics[width=0.99\textwidth]{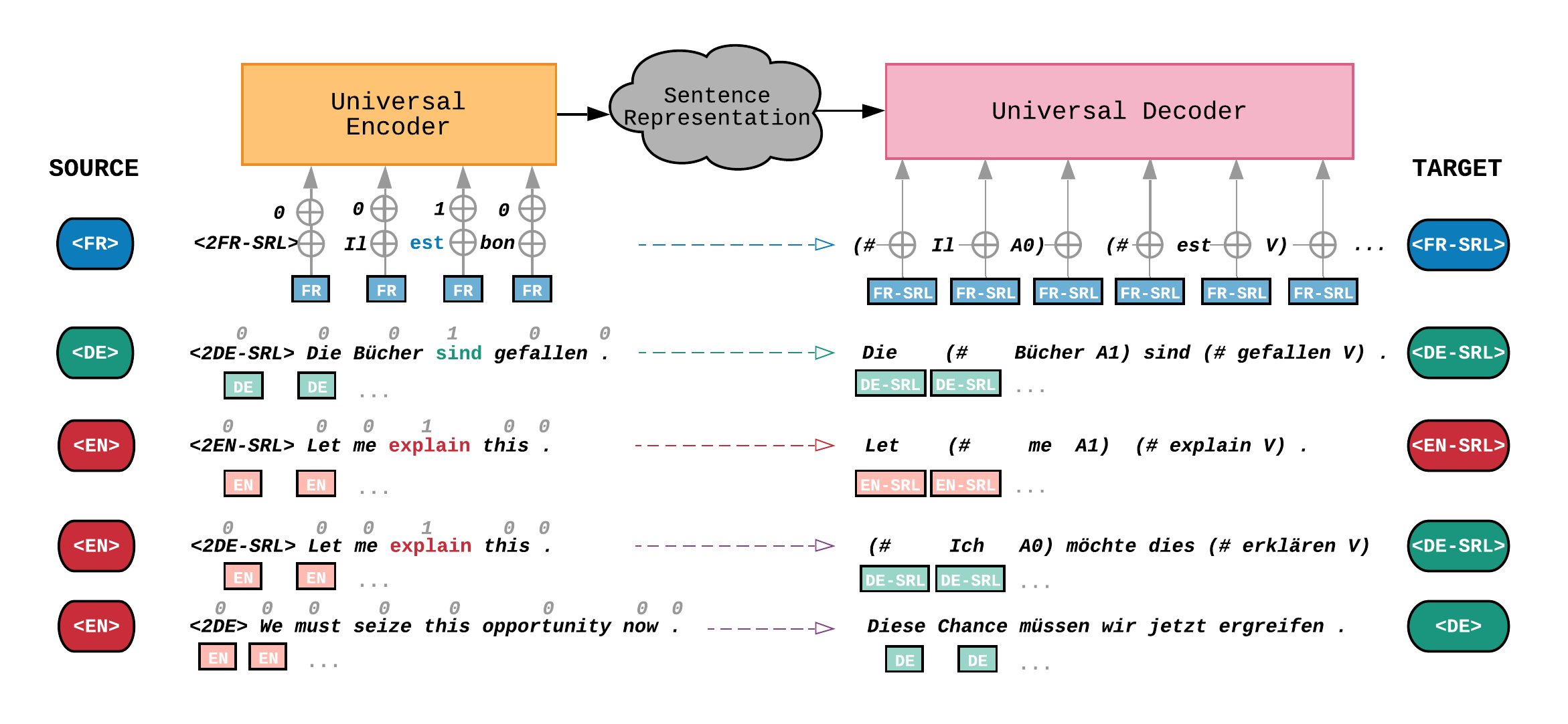}
  \caption{A universal structure-inducing Enc-Dec model with copying and sharing of parameters across languages that jointly translates and labels sentences. It can also be trained cross-lingually and be augmented with classic machine translation data (as shown in the two bottom rows of the figure).}
\label{fig:model_multi}
\end{figure*}

We define the SRL task as a sequence transduction problem: given an input sequence of tokens $X=x_{1}, ..., x_{i}$, the system is tasked to generate a sequence $Y=y_{1}, ..., y_{j}$ consisting of words interleaved with SRL annotations. Defining the task in this fashion allows $X$ and $Y$ to be of different lengths and therefore target sequences may also contain word tokens of different languages if desired. This means that we could train an Enc-Dec model that learns not only to label a sentence, but to jointly translate it while applying SRL annotations directly to the target language. Moreover, following conceptually the multilingual Enc-Dec model proposed by \citet{Johnson16-NMT}, we can train a single model that allows for joint training with multiple language pairs while sharing parameters among them. We apply a similar joint multilingual learning method to produce structured output sequences in the form of translations enriched with SRL annotations on the (lower-resource) target language (cf.\ Figure \ref{fig:model_multi}). We will apply this universal structure-inducing Enc-Dec model to the Semantic Role Labeling task, and show that it can be deployed in three different settings:

\begin{figure}
\centering
  \includegraphics[width=0.49\textwidth]{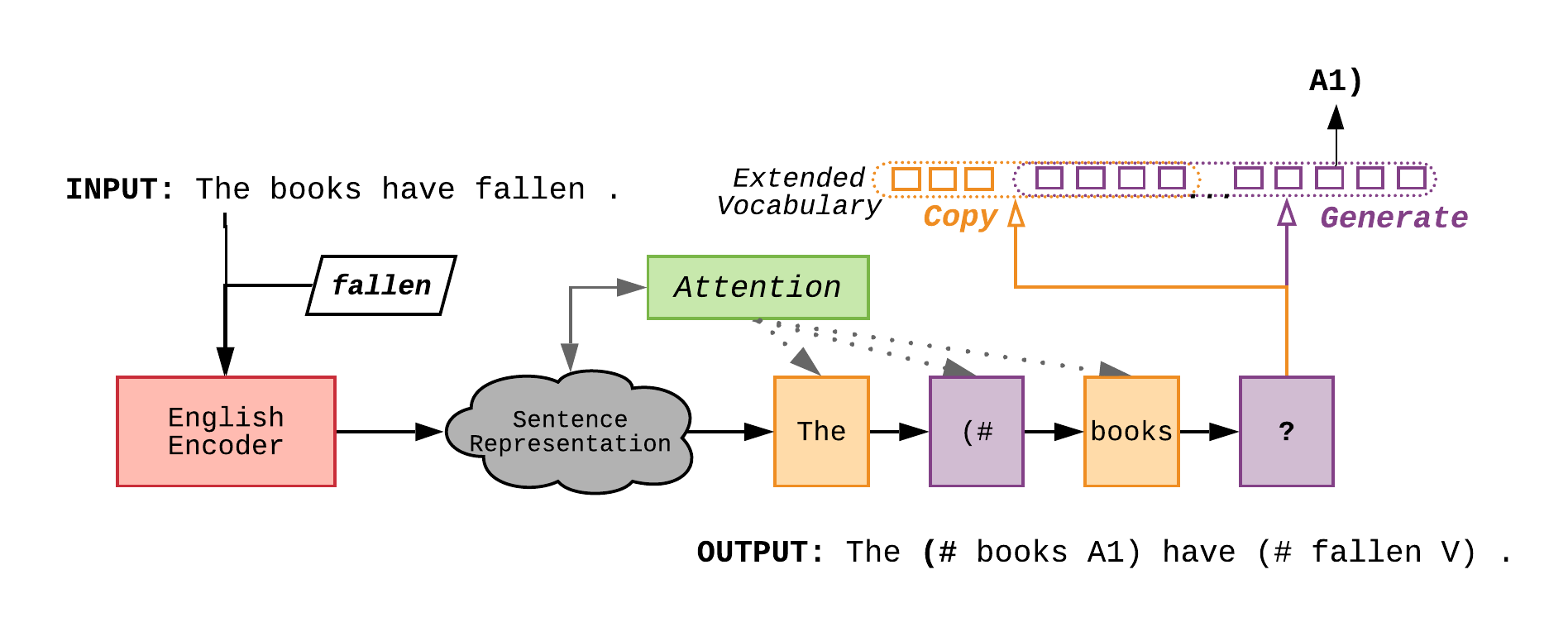}
  \caption{Monolingual Enc-Dec model for SRL with copying
  \cite{Daza18-SRL}. 
  We generalize this architecture to multilingual and cross-lingual SRL.}
\label{fig:model_mono}
\end{figure}

i) \textbf{monolingual}: encode a sentence in a given language and learn to decode a labeled sequence by reproducing the source words and inserting the appropriate structure-indicating labels in the output (cf.\ Figure \ref{fig:model_mono}). A copying mechanism \cite{Gu16-S2S} allows this model to reproduce the input sentence as faithfully as possible.

ii) \textbf{one-to-one multilingual}: train a single, joint model to generate $n$ different structure-enriched target languages given inputs in the same language. For example: Labeled English (\textit{EN-SRL}) given an \textit{EN} sentence or Labeled German (\textit{DE-SRL}) given a \textit{DE} sentence. This multilingual model still relies on copying to relate each labeled output sentence to its corresponding input counterpart. However, unlike (i), it has the advantage of sharing parameters among languages.

iii) \textbf{cross-lingual}: generate outputs in $n$ different target languages given inputs in $m$ different source languages, for example: Labeled German (\textit{DE-SRL}) and Labeled French (\textit{FR-SRL}) given an \textit{EN} sentence (see Figure \ref{fig:model_multi}). In this setting, we do not restrict the model to \textit{copy} words from the source sentence but train it to \textit{translate} them.

In Section \ref{sec:mono-model} we describe how the basic Enc-Dec model for SRL is constructed and in Section \ref{sec:multi-model} we describe the additional components that allow us to generalize this architecture to the one-to-one multilingual and cross-lingual scenarios.

\subsection{Encoder-Decoder Architecture}
\label{sec:mono-model}

We reimplement and extend the Enc-Dec model with attention \cite{Bahdanau15-S2S} and copying \cite{Gu16-S2S} mechanisms for SRL proposed by \citet{Daza18-SRL}. This model encodes the source sentence and decodes the input sequence of words (in the same language) interleaved with SRL labels. 

\textbf{Data Representation.}
Similar to other prior work \cite{liu-etal-2018-discourse} and our own \cite{Daza18-SRL}, we linearize the SRL structure in order to process it as a sequence of symbols suitable for the Enc-Dec architecture. We restrict ourselves to argument identification and labeling of one predicate at a time. We feed the gold predicate in training and inference, and process each sentence as many times as it has predicates. An opening bracket \textit{(\#} indicates the start of a labeled-argument region; a closing labeled bracket, e.g.\ \textit{A0)}, indicates the ending and the tag of the labeled region (see Figure \ref{fig:model_mono}).

\textbf{Vocabulary.} 
We define a shared vocabulary consisting of all source and target words $\mathcal{V}=\left\{v_{1},...,v_{N}\right\}\cup \left\{UNK\right\}$ and the SRL labels $\mathcal{L}=\left\{l_{1},...,l_{M}\right\}$. In addition, we employ a per-instance extension set $\mathcal{X} = \left \{x_{1}...,x_{T_{x}} \right \} $ containing all words from the source sequence. Our final vocabulary is $\mathcal{V}\cup \mathcal{L}\cup \mathcal{X}$. 

\textbf{Encoder.}
In our prior work \cite{Daza18-SRL} we used a 2-layer BiLSTM as encoder. In this paper, we adopt the Deep BiLSTM Encoder from \citet{He17-SRL} which has been shown to work well for SRL models. Again following \citet{He17-SRL}, we define the encoder input vector $x_{i}$ as the concatenation of a word embedding $w_{i}$ and a binary predicate-feature embedding $p_{i}$ indicating at each time-step whether the current word is a predicate or not \footnote{These two additions already show improvements compared to the reported results in \citet{Daza18-SRL}.}. The encoder outputs a series of hidden states $h_1, ...,  h_{T_{x}}$ representing each token. We refer to this series of states as \textbf{H}.

\textbf{Attention.} To improve the access to the source sentence representation, we include the attention mechanism proposed by \citet{Bahdanau15-S2S}, which computes a context vector at each time step {\it t} based on \textbf{H} and the current decoder state.

\textbf{Decoder.} We use a single-layer Decoder with LSTM cells \cite{Hochreiter:LSTM} and a copying mechanism. It emits an output token $y_{t}$ from a learned score $\psi_{g}$ over the vocabulary at each time step {\it t} given its state $s_{t}$, the previous output token $y_{t-1}$, and the attention context vector $c_{t}$.
In addition, a copying score $\psi_{c}$ is calculated. The decoder learns from these scores when to generate a new token and when to copy from the encoded hidden states \textbf{H}. Formally we compute the scores as:

\vspace*{-2ex}
\begin{equation}
\begin{split}
\psi_{g}(y_{t} = v_{i}) = W_{o}[s_{t};c_{t}], &~~~ v_{i}~ \epsilon ~\mathcal{V} \cup \mathcal{L}\\ 
\psi_{c}(y_{t} = x_{j}) = \sigma(h_{j}^{T}W_{c})s_{t}, &~~ x_{j} ~ \epsilon ~\mathcal{X}
\end{split}
\end{equation}

where $W_{o} \epsilon \mathbb{R}^{N\times 2d_{s}}$ and $W_{c} \epsilon \mathbb{R}^{d_{h}\times d_{s}}$ are learnable parameters and $s_{t}$, $c_{t}$ are the current decoder state and context vector, respectively.
These scores are used to compute two distributions: one for the likelihood of copying ($\mathbf{c}$) $y_{t}$ and another for the likelihood of generating ($\mathbf{g}$) $y_{t}$. Formally:
\begin{multline}\label{dec_prob}
p(y_{t} | s_{t},y_{t-1},c_{t},\mathbf{H}) = p(y_{t},\mathbf{g} | s_{t},y_{t-1},c_{t}) + \\ p(y_{t},\mathbf{c}| s_{t},y_{t-1},\mathbf{H})
\end{multline}

The two distributions are then normalized by a final {\it softmax} layer from which we compute a joint likelihood of $y_{t}$ and choose the token with the highest score within this joint likelihood. 

\subsection{Multilingual Extensions}
\label{sec:multi-model}

We generalize the monolingual Enc-Dec model for SRL to a multilingual SRL system by adding two main components:

\textbf{Translation Token.}
Like \citet{Johnson16-NMT}, we prefix the source sequence with a special token that indicates the expected language of the target sequence. If the source is in \textit{EN} and the target is a German sentence with SRL labels, the source sentence will be preceded by the token $<$\textit{2DE-SRL}$>$.

\textbf{Language Indicator Embeddings.} We want the model to profit from the common role label inventory used across languages, yet at the same time there are subtle differences in role labeling and how roles are linguistically marked in the different languages\footnote{e.g. the role A2 (Beneficiary) can be PP in \textit{EN} and \textit{FR}, but dative NP in \textit{DE} (DativeNP)}.
Hence, we define \textit{N} different language indicators (e.g., \textit{FR, DE}) and represent each of them with a randomly initialized language indicator vector that we tune during training. The model can use these language indicator embedding vectors to leverage language-specific properties when generating SRL annotations. Also, by using these embeddings in the decoder, we can help it to stay consistent regarding the language it generates\footnote{\citet{Johnson16-NMT} only use a translation token, but our training data is significantly smaller than theirs.}. 

Thus, in all multilingual settings, at each time step $t$ we feed the \textbf{Encoder}  with a concatenation of the previous encoder state $h_{t-1}$, the word embedding $w_{t}$ of the current token, the embedded predicate indicator $p_{t}$ and the language indicator embedding $l_{t}$. The Encoder state update is defined as:

\begin{equation}
h_{t}= LSTM([h_{t-1};w_{t};p_{t};l_{t}])
\end{equation}

Likewise, on the \textbf{Decoder} side we concatenate the representations for both word tokens and label tokens with the language indicator vector to produce tokens in a specific language. For SRL-labeled output sentences the indicator token for the language embedding is \textit{DE-SRL, FR-SRL, ...} depending on the target language. Formally, at each time step the decoder updates its state by taking into account the previous decoder state $s_{t-1}$, the previous generated token\footnote{During training we use teacher forcing, feeding the gold target token instead of the previously generated token} $y_{t-1}$, the language indicator embedding $l_{t-1}$ and the attention context vector $c_{t}$:  

\begin{equation}
s_{t}= LSTM([s_{t-1};y_{t-1};l_{t-1};c_{t}])
\end{equation}

\section{Data}
\label{sec:data}

\subsection{SRL Monolingual Datasets}
Two labeling schemes have been established for PropBank SRL: span-based and dependency-based. In the former, arguments are characterized as word-spans. This scheme was introduced in the CoNLL-05 Shared-task \cite{Carreras05-SRL} and is available only for English. In the dependency-based SRL format, only the syntactic heads of arguments are labeled. This format was defined in the CoNLL-09 Shared task \cite{Hajic09-SRL}, which includes SRL labeled data for seven languages.\footnote{Note that CoNLL-09 is not a parallel corpus. All data was annotated independently and later ported to CoNLL.} 

We will use span-based and dependency-based data for English to benchmark the monolingual system. For the multilingual experiments, we use the dependency-based annotations, given that there is labeled data available in different languages on this format. Specifically, we use the English and German portions of CoNLL-09 and the automatically annotated French SRL corpus of \citet{vanDerPlas11-Proj} for training and the human-labeled sentences from \citet{vanDerPlas10-Proj} for testing. Both corpora contain a similar label set as the English PropBank\footnote{French data was directly annotated using the English labelset but German CoNLL-09 contains additional core labels A5-A9 and does not contain \textit{AM-} modifier labels}. For statistics on the size of the datasets see Table \ref{Data-Mono}.

\begin{table}[t!]
\centering
\resizebox{0.48\textwidth}{!}{
\begin{tabular}{@{}|ll|ll|l|@{}}
\hline
Mono- & Language      & \multicolumn{2}{c}{Train} &   Test\\
lingual &  & \# Sents & w/ 1-Pred &\ w/ 1-Pred\\ 
\hline
CoNLL-05          & EN {[}Span{]} & 75,187       & 94,497    &  5,476    \\
CoNLL-09          & EN {[}Dep{]} & 39,279       & 180,446   & 10,626  \\
CoNLL-09          & DE {[}Dep{]} & 36,020       & 39,138    & 2,044 \\
v.d. Plas  & FR {[}Dep{]} & 40,075       & 73,094          & 2,036  \\ 
\hline
\end{tabular}
}
\caption{Train and Test Data for Monolingual Models. We show the original number of sentences and the size of the "expanded" data with one copy per predicate.}
\label{Data-Mono}
\end{table}
\begin{table}[t!]
\centering
\resizebox{0.38\textwidth}{!}{
\begin{tabular}{|l|l|}
\hline
Cross-lingual Model       & \# Sentences \\ \hline
EN - DE-SRL (Akbik, 2015) & 63,397       \\
EN - FR-SRL (Akbik, 2015) & 40,827       \\
EN - FR (UN)              & 100,000      \\
EN - DE (Europarl)        & 100,000      \\ \hline
\end{tabular}
}
\caption{Data used for Cross-lingual Models: From the SRL parallel data available we take $90\%$ for training and use the rest as a \textit{Dev} set for our experiments.  We add the non-labeled data (from UN and Europarl) during training to enforce translation knowledge.}
\label{Data-Cross}
\end{table}

\subsection{Datasets for Cross-lingual SRL}
\label{subsec:cross_data}
We use the dependency-based labeled German and French SRL corpus from \citet{Akbik15-Proj} which was produced via annotation projection. These sentences are already pre-filtered to ensure that the predicate sense of the source predicate is preserved in the target sentence. Since the role labels are projected from automatically PropBank-parsed English sentences, all languages share the same label set. The underlying corpus for this dataset is composed of Machine Translation (MT) parallel corpora: Europarl \cite{Koehn05-Data} for \textit{EN-DE} (about 63K sents), and UN \cite{Ziemski16-Data} for \textit{EN-FR} (about 40K sents).

Since we only had access to the labeled sentences (target-side), we constructed our parallel training pairs \textit{EN} to \textit{FR-SRL} and \textit{EN} to \textit{DE-SRL} by finding the original source English counterparts. We use Flair \cite{akbik18-Data} to predict PropBank frames on the English source sentences and find the alignment to the labeled predicate on the target side using fast-align \cite{dyer13-Data}. 

In addition to the parallel SRL-labeled data, we choose a subset of 100K parallel (non-labeled) sentences for each language pair from the mentioned MT datasets (Europarl and UN corpora) to improve the translation quality of the model, we use $90\%$ for training and the rest as a development set. The data is summarized in Table \ref{Data-Cross}.

\section{Experiments and Results}
\label{sec:experiments}

\subsection{General Settings}
We use the AllenNLP \cite{Gardner17-AllenNLP} Enc-Dec model as a basis for our implementation. Our model is trained to minimize the negative log-likelihood of the next token. Hyperparameters and model sizes are provided in Supplement A.1. We use pre-trained word embeddings (fine-tuned during training) for the 3 languages: GloVe \cite{Pennington14glove-Exps} for \textit{EN} and the pre-trained vectors from \citet{grave18-Exps} for \textit{FR} and \textit{DE}. We also train versions with contextual word representations: pre-trained English 1024-dimensional ELMo \cite{Peters18-elmo-exps} and multilingual 768-dimensional BERT-small \cite{Devlin18-bert-exps} representations.

\subsection{Monolingual Experiments and Results}
We train three separate monolingual versions for \textit{EN}, \textit{DE} and \textit{FR}. We first benchmark our system against a wide variety of English models (span- and dependency-based) that perform the role classification task with gold predicates to show that our labeling performance is competitive with the existent SOTA neural models for English. This is shown in Table \ref{table:eval-eng}. The performance of \textit{DE} and \textit{FR} is shown in Table \ref{SoA-multilang} where we compare all monolingual systems for the three languages (top half), against the one-to-one multilingual versions (bottom half). Results for \textit{EN} show that the Enc-Dec architecture is competitive with the GloVe-based models (although still 4 F1 points below SOTA in most cases), however it benefits more from ELMo, achieving SOTA results for span-based and dependency-based SRL.

\subsection{Multilingual Experiments and Results}
\label{subsec:multi-exps}
We train a single multilingual model with the concatenation of the training data for the three languages \textit{EN}, \textit{DE} and \textit{FR} that we previously used on the monolingual experiments. We use a common vocabulary for the three languages and keep all tokens that occur more than 5 times in the combined dataset. We train the model with batches containing instances randomly chosen from the individual languages (this means that each batch might contain examples from different language pairs). 

Multilingual training yields improvement on the three languages studied in this paper when compared to our monolingual baselines, particularly for German, which shows more than 6 points (F1) of improvement. In addition, we compare with the polyglot SRL system of \citet{Mulcaire18-SRL} (which also leverages data from multiple languages during training), obtaining better results for English using GloVe. We then show that adding contextual representations to our model results in bigger improvements across the board.

\begin{table}[]
\centering
\resizebox{0.48\textwidth}{!}{

\begin{tabular}{@{}l@{~~}l@{~~}l@{~}l@{~}l@{~}l@{~}l@{}}
Type            & Model        & Word               & \multicolumn{2}{c}{CoNLL-05} & \multicolumn{2}{c@{}}{CoNLL-09}  \\
                &           & Repres.               & \multicolumn{1}{c@{}}{WSJ} &  \multicolumn{1}{c@{}}{OOD} & \multicolumn{1}{c@{}}{WSJ} & \multicolumn{1}{c@{}}{OOD} 
   \\\hline
\multirow{7}{*}{\begin{sideways}Span SRL\end{sideways}}               & He 2017                     & GloVe                  & 84.6          & 73.6              & -            & -                \\
                & Daza 2018             & GloVe                  & 79.2          & 68.4              & -            & -                \\
                & He, 2018                     & ELMo                   & 83.9          & 73.7              & -            & -                \\
     & Tan, 2018                    & GloVe                  & 84.8          & 74.1              & -            & -                \\
                & Strubell 18 {[}LISA{]}    & GloVe                  & 84.6          & 74.5              & -            & -                \\
                & Strubell 18 {[}LISA*{]}   & ELMo                   & 86.5          & 78                & -            & -                \\
                & Ouchi 2018  & ELMo                   & \textbf{88.5}          & 79.6              & -            & -                \\ \hline
\multirow{3}{*}{\begin{sideways}Dep\end{sideways}}                 \multirow{3}{*}{\begin{sideways}SRL\end{sideways}} & Roth 2016                   & DPE*         & -             & -                 & 87.7         & 76.1             \\
             & Marcheggiani 2017           & Dyer*  & -             & -                 & 87.7         & 77.7             \\
                & Cai et al 2018              & GloVe                  & -             & -                 & 89.6         & 79               \\ \hline
\multirow{4}{*}{\begin{sideways}Dep and\end{sideways}}             \multirow{4}{*}{\begin{sideways}Span SRL\end{sideways}}     & FitzGerald 2015             & GloVe                  & 80.3          & 72.2              & 87.8         & 75.5             \\
                & Li 2019               & ELMo                   & 87.7          & -                 & 90.4         & -                \\
                & Ours {[}Mono{]}              & GloVe                  & 80.4          & 70.5              & 85.5         & 75.7             \\
                & Ours {[}Mono{]}              & ELMo                   & 88.3          & \textbf{80.9}              & \textbf{90.8}         & \textbf{84.1}     \\ \hline   
\end{tabular}

}
\caption{CoNLL-09 and CoNLL-05 Test Sets for English. Our model with ELMo shows SOTA performance on both types of SRL. LISA* only reports ELMo with predicted predicates; DPE*: dependency path embeddings; Dyer*: Dyer et al. 2015.}
\label{table:eval-eng}

\end{table}

\begin{table}[t!]
\centering
\resizebox{0.48\textwidth}{!}{
\begin{tabular}{@{}l|l@{~~}l@{~~}l@{}}
Model                              & EN-Test & DE-Test & FR-Test \\ \hline
SOTA models*                       & 90.4    & 80.1    & 73      \\
Ours-EN {[}Mono + GloVe{]}         & 85.5    & -       & -       \\
Ours-DE {[}Mono + GloVe{]}         & -       & 61.9    & -       \\
Ours-FR {[}Mono + GloVe{]}         & -       & -       & 70.3    \\ \hline
Mulcaire 2018 {[}Multi + GloVe{]} & 86.5    & 69.9    & -       \\
Ours {[}Multi + GloVe{]}           & 87      & 68.2    & 70.5    \\
Ours {[}Multi + ELMo{]}            & 91.1    & 75.7    & 70.7    \\
Ours {[}Multi + BERT{]}            & 89.7    & 77.2    & 72.4   \\ \hline
\end{tabular}
}
\caption{F1 scores for role labeling on dependency-based SRL data. EN and DE Tests: CoNLL-09; FR-Test: \citet{vanDerPlas11-Proj}. State of the art (SOTA) models$^*$ are: \citet{Cai18-SRL} [GloVe] for EN, \citet{Roth16-REL} [Dependency-path Embeddings] for DE and \citet{vanDerPlas14-Cross} [Non-neural] for FR, respectively.
}
\label{SoA-multilang}
\end{table}

\subsection{Cross-Lingual Experiments and Results}
\label{subsec:cross-exps}

\paragraph{Training.} 
After validating the robustness of our architecture when handling different languages at the same time, we now train a cross-lingual SRL version. This setting differs from the previous two because the model needs to learn two tasks: besides generating appropriate SRL labels, it needs to translate from source into a target language. To do so, we train a single model using the concatenation of the parallel datasets listed in Table \ref{Data-Cross} and described in Section \ref{subsec:cross_data}. We further include Machine Translation (MT) data to reinforce the translation knowledge of the model, so that it can generate fluent (labeled) target sentences. As in the multilingual experiments, we train the model with alternating batches of instances randomly chosen from the individual language pairs. Note that the amount of MT data that we can add is restricted: the labeled multilingual data is relatively small and labeling performance suffers when the MT data gets too dominant.

\paragraph{Evaluating Cross-lingual SRL.}
As in classical MT, evaluation is difficult, since the system outputs will approximate a target reference but will never be guaranteed to match it. Hence in this setting we do not have a proper gold standard to evaluate the labeled outputs, since we are generating labeled target sentences from scratch. Similar to MT research, we apply BLEU score \cite{papineni-02-bleu} to measure the closeness of the outputs against our \textit{Dev} Set. 

The upper part of Table \ref{CrossBLEU} compares the scores of two versions of the Enc-Dec model trained on the cross-lingual data from Table \ref{Data-Cross} systems, one using GloVe embeddings and the second using BERT, respectively. To better distinguish translation vs. labeling quality, we compute BLEU scores for the system outputs against labeled reference sentences in three different ways: on \textit{words only}, \textit{labels only}, and on \textit{full labeled sequences} (both word and label outputs). We see that the prediction of words is similar in the two languages, but labeling is more difficult for \textit{DE} than for \textit{FR} for both systems. Also we observe that adding multilingual BERT is very helpful to obtain even more fluent and correct labeled outputs (according to BLEU) resulting in ca. +9 points in German and +5 in French on the full sequences. This is very important given that we have a small training set compared to classic NMT scenarios. 

The bottom part of Table \ref{CrossBLEU} shows the scores when restricting the evaluation to sentences with score $\geq$ 10. We observed that this threshold\footnote{We tried with thresholds of 5, 10, 20 and 30.} is a good trade-off in both the amount of kept sentences (above the threshold) and average BLEU score increase (presumably sentence quality). By keeping only the filtered subset of sentences we get an improvement on average of  approx. 10 BLEU points on the full sequences (\textit{F-Seq}), and almost double the score for \textit{labels only}. This holds for GloVe and BERT versions on both languages.

\begin{table}[t!]
\centering
\resizebox{0.48\textwidth}{!}{
\begin{tabular}{@{}l|l@{~~}l@{~~}l|l@{~~}l@{~~}l|@{}}
                                   & \multicolumn{3}{c|}{\textbf{German}}                        & \multicolumn{3}{c|}{\textbf{French}}                         \\ \hline
\textbf{Model {[}Filter{]}}        & \textbf{F-Seq} & \textbf{Word} & \textbf{Label} & \textbf{F-Seq} & \textbf{Word} & \textbf{Label} \\ \hline
XL-GloVe {[}All{]}              & 18.86             & 17.17          & 25.52           & 28.99             & 17.36          & 32.76           \\
XL-BERT {[}All{]}               & 27.22             & 27.36          & 29.59           & 33.59             & 22.48          & 37.17           \\ \hline
XL-GloVe {[}$\geq{10}${]} & 30.58             & 36.71          & 51.68           & 38.99             & 43.79          & 61.73           \\
XL-BERT {[}$\geq{10}${]}  & 36.95             & 41.36          & 55.73           & 42.66             & 46.52          & 65.32          
\end{tabular}
}
\caption{Cross-lingual (XL) system results using BLEU score on individual languages inside the \textit{Dev} set. We compute BLEU on labeled sequences (F-Seq), and separately for words and only labels. We also show scores when pre-filtering on F-Seq with BLEU $\geq{10}$.
}
\label{CrossBLEU}
\end{table}

\paragraph{Output Filtering and Data Generation.}
We use our cross-lingual model as a labeled data generator by applying it on \textit{EN} sentences from Europarl (100K) and UN corpora (100K)\footnote{Note that these are taken from a different subset than the parallel sentences used during training.} and let the model predict \textit{DE-SRL} and \textit{FR-SRL} as target languages. This results in \textit{unseen} German and French labeled sentences. Since we cannot guarantee that the generated sentences preserve the source predicate meaning, we filter all outputs by keeping only those that come close to the original sentence meaning. We approximate this by back-translating the generated outputs (stripping the labels and keeping only the words) using the pre-trained \textit{DE-EN} model from OpenNMT \cite{Klein-17-opennmt-exps}. 

We compare the back-translations to the sentences that we originally presented to the system and, using the previously described filtering heuristic, we keep only those whose BLEU score is equal or greater than 10. The logic behind this is that if the back-translation is close enough to the source, the generated sentence preserves a fair amount of the original 
sentence meaning\footnote{BLEU score is used as a naive approach to avoid excessively noisy data but we could also develop, for example, a semantic similarity metric to also keep sentences that are close enough to the original predicate sense meaning.}. 
With this strategy, after applying the \textit{BLEU filter}, we end up with a parallel dataset of 44K generated sentences for \textit{(EN, DE-SRL)} and 32K for (\textit{EN}, \textit{FR-SRL}). In the next section we show more detailed evaluation measures of the system outputs, focusing on the filtered dataset that we just described.



\subsection{Cross-Lingual Detailed Evaluation}
\label{sec:xl-eval}
We are aware that BLEU score gives only a rough estimate of the actual quality of the outputs, therefore we propose to measure the performance of our system in two more detailed evaluation settings: (i) a small-scale \textbf{human evaluation} where we evaluate the assigned SRL labels against 226 sentences that were manually judged and annotated to give an estimation of the quality of the generated data, (ii) an \textbf{extrinsic evaluation} using labeled sentences generated by our system to augment the training set for a resource-poor language. We conduct the extrinsic evaluation on German and French and the manual evaluation only on the German data, which proved to be the more challenging language compared to French.

\subsubsection{Human Evaluation}
To provide an in-depth quality assessment of the generated sentences, we create a small-scale gold standard consisting of 226 sentences. To select a representative sample from our newly generated labeled sentences,\footnote{i.e., the generated sentences for which we measured a BLEU score $\geq$ 10 against the source using back-translation.} we analyze the distribution of labels in the data and apply stratified sampling to cover as many predicates as possible and as many role label variants as possible. We judge these sentences on the quality of the generated language and annotate them with PropBank roles.

\paragraph{SRL Gold Standard.} As we are lacking trained PropBank annotators, we mimic the question-based role annotation method of \citet{he-15-QA-SRL}, who constructed QA pairs in order to label the predicate-argument structure of verbs. The annotation involves several subtasks: The first is to generate questions targeting a specific verb in a sentence and to mark as answers a subset of words from the same sentence. The next subtask is to choose the head word of each selected subset and to assign a PropBank label to this head according to a table that correlates WH-phrases with the most likely label.\footnote{We provide this correlation table and the full annotation guidelines in appendix A.2 in the Supplement.} 

We ask two linguistically trained annotators to perform the whole task independently and compute Krippendorff's Alpha \cite{Krippendorff-alpha-80} on the role labels, which results in an inter-annotator agreement score of 82.83. We resolved conflicting annotations through discussion among the annotators. The resulting gold standard contains 737 annotated roles. Notably, the most prominent roles (as in the CoNLL datasets) are A0 and A1 which are normally related to the agent and the patient in sentences, but the annotated data also includes modifier roles such as temporal, modal, discourse markers, among others\footnote{The label distribution is given in the Supplement, A.3.}. 

\paragraph{Translation Quality.} 
We ask two different annotators to score each output sentence (they see only the words, not the labels) on a scale of 1-5 for \textit{Quality} (1: `is completely ungrammatical'; 5: `is perfectly grammatical') and for \textit{Naturalness} (1: `The sentence is not what a native speaker would write'; 5: `The sentence could have been written by a native speaker'). We obtain a high average score of 4.4 for \textit{Quality} and 4.2 for \textit{Naturalness}.

\paragraph{SRL Performance on Gold Standard.}
We use our human-annotated sentences to measure the automatic labeling performance of our cross-lingual SRL model which we call XL-BERT). We obtain 73.21 F1 score (73.33 precision, 73.1 recall). We also measure the performance of the ZAP label projection system of \citet{akbik18-zap} on this data (we only consider arguments of the predicates that were annotated). ZAP obtains a low F1 score of 56.03 (42.65 precision, 81.7 recall). Thus, XL-BERT shows much better, and more precise results compared to this baseline and achieves overall very acceptable and stable labeling quality. 
This shows that the joint translation-labeling task is successful. ZAP, by contrast, shows more unstable results, which might be due to word alignment noise. Although we train on such data, our model can also loose some of this noise, given that the same model is trained to produce more than one labeled language, namely \textit{FR-SRL} and \textit{DE-SRL}.

\subsubsection{Extrinsic Task: Data Augmentation}
Finally, we augment the training sets of our two resource-poor languages \textit{DE} and \textit{FR}, in portions of 10K until we cover the complete generated data. We compare the increase in F1 score when training models with different amounts of additional data. We also add a comparison of the improvement achieved when adding the same amount of sentences produced by the labeled projection method of \citet{Akbik15-Proj}. 
We see in Table \ref{AugmentedData} that adding our German data shows improvement in F1 score, despite the fact that the CoNLL-09 label scheme has arguments not seen in our training data (namely A5-A9). Presumably we see this improvement because the frequency of the major roles is more prominent. In the case of French, we don't see significant improvement, however also here the addition of projected data shows a similar trend.

\begin{table}[t!]
\centering
\resizebox{0.48\textwidth}{!}{
\begin{tabular}{@{}lll@{}}
Model  + Training Data    & Data Size & F1 Test \\ \hline
DE {[}Mono{]}~~ \textit{(Original)}  & 39K       & 61.9    \\
DE {[}Mono{]} + \textit{LabelProj} & 83K       & 62.37    \\
DE {[}Mono{]} + OurGen10K & 49K       & 62.4    \\
DE {[}Mono{]} + OurGen20K & 59K       & 62.46   \\
DE {[}Mono{]} + OurGen30K & 69K       & 62.81   \\
\textbf{DE {[}Mono{]} + OurGenALL} & \textbf{83K}       & \textbf{63.57}   \\ \hline
FR {[}Mono{]}~~ \textit{(Original)}  & 73K       & 70.3    \\
FR {[}Mono{]} + \textit{LabelProj} & 105K      & 70.45   \\
FR {[}Mono{]} + OurGen10K & 83K       & 70.33   \\
\textbf{FR {[}Mono{]} + OurGen20K} & \textbf{93K}       &\textbf{70.52}  \\
FR {[}Mono{]} + OurGenALL & 105K      & 70.39  
\end{tabular}
}
\caption{We retrain the monolingual systems \textit{DE, FR} using the original training sets (BL: \textit{Original}) shown in Table \ref{Data-Mono} and inject our generated data in different sizes. We also compare to the stronger baseline \textit{LabelProj} where we add data created by label projection  \cite{Akbik15-Proj}.}
\label{AugmentedData}
\end{table}

\section{Related Work}

\textbf{Encoder-Decoder Models}. A wide range of NMT models are based on the Encoder-Decoder approach \cite{Sutskever14-REL} with attention mechanism \cite{Bahdanau15-S2S, Luong15-REL}. More recent architectures \cite{Zoph16-NMT, Firat16-multiNMT} show that training with multiple languages performs better than one-to-one NMT. Multilingual models have also been trained to perform Zero-shot translation \cite{Johnson16-NMT, Firat16-zeroNMT}. The Enc-Dec approach has 
been tested in many tasks that can be formulated as a sequence transduction problem:
syntactic parsing \cite{Vinyals14-REL}, AMR and Semantic Parsing \cite{Konstas17-REL,Dong16-REL} and SRL \cite{Daza18-SRL}. The most similar approach to ours is \citet{zhang17-REL}, who propose a cross-lingual Enc-Dec that produces OpenIE-annotated English given a Chinese sentence. However, their setup is easier than ours since they have a reliable labeler on the target side, facilitating the generation of more training data unlike us who are interested in labeling the resource-poor language.

\textbf{Cross-lingual Annotation Projection.} A common approach to address the lack of annotations is projecting labels from English to a lower-resource language of interest. This has shown good results in the transfer of semantic information to target languages. \citet{KozhevnikovP13-REL} propose an unsupervised method to transfer SRL labels to another language by training on the source side and using shared feature representations for predicting on the target side. \citet{Pado09-Proj} project FrameNet \cite{Baker98-SRL} SRL labels by searching for the best alignment in source and target constituent trees, defining label transfer as an optimization problem in a bipartite graph. \citet{vanDerPlas11-Proj} use intersective word alignments between English and French with additional filtering heuristics to determine whether a PropBank label should be transferred and then use this to train a joint syntactic-semantic parser for both languages. \citet{Akbik15-Proj} proposes a higher-confidence projection by first creating a system with high precision and low recall and then using a bootstrap approach to augment the labeled data. 

Separately, \citet{MEANTIME:2016} generated a multilingual event and time parallel corpus including SRL annotations. Their corpus was manually annotated on the English side and automatically projected to Italian, Spanish, and Dutch based on the manual alignment of the annotated elements. Unfortunately, the authors do not report the performance of the SRL task, making it difficult for us to use their data for benchmarking.

\textbf{Semantic Role Labeling.}
Span-based SRL only exists on English data \cite{Zhou15-SRL, He18-SRL, Strubell18-REL, Ouchi18-REL}. Dependency-based SRL models such as \cite{March17-2-REL, Cai18-SRL, Li19-REL} are the state-of-the-art for English. For French, we compare against \citet{vanDerPlas14-Cross} since we did not find more recent work for that language. \citet{Roth16-REL} show a model based on dependency path embeddings that achieved SOTA in English and German. The Polyglot SRL model of \citet{Mulcaire18-SRL} shows some improvement over monolingual baselines when aggregating all multilingual data available from CoNLL-09, while more refined integration did not show further improvement. Their system does not perform better than our multilingual models for English and German.

\section{Conclusions}

We presented the first cross-lingual SRL system that translates a sentence and concurrently labels it with PropBank roles. The proposed Enc-Dec architecture is flexible: as a \textit{monolingual} system the model achieves SOTA for English PropBank role labeling, the \textit{multilingual} SRL system shows that joining multiple languages improves SRL performance over the monolingual baselines, and a \textit{cross-lingual} system can be used to generate SRL-labeled data for lower-resource languages. Evaluation of the cross-lingual system shows that the quality-filtered sentences are highly grammatical and natural, and that the generated PropBank labels can be more precise than label projection. Using our labeled data beats a label projection baseline when using it to augment the training set of a lower-resource language. 

An advantage of our proposed model is that it does not need parallel data at inference time. Our current model can possibly be further improved by adding more automatically generated data in the data augmentation scenario, or by targeted selection in an active learning setting. Current limitations of the system may be alleviated by pre-training the model to acquire better translation knowledge from larger training data, and by developing more refined filtering methods. 

In future work we also aim to make the system more flexible, by extending it to few-shot or zero-shot learning, to alleviate the need for an initial big annotated set, and thus to be able to generate SRL data for truly resource-poor languages. Further challenges for this novel architecture are to extend it to joint predicate and role labeling for more than one predicate at a time.

\section*{Acknowledgements}
We thank the reviewers for their insightful comments. This research is funded by the Leibniz ScienceCampus Empirical Linguistics
\& Computational Language Modeling, supported by Leibniz Association grant no. SAS2015-IDS-LWC and by the Ministry of Science, Research, and Art of Baden-Wurttemberg. We thank NVIDIA Corporation for donating GPUs used in this research. We are grateful to our annotators and to \'{E}va M\'{u}jdricza-Maydt for her assistance with the human evaluation setup.

\bibliography{Main}
\bibliographystyle{acl_natbib}

\appendix

\end{document}